# A Survey of Named Entity Recognition in Assamese and other Indian Languages

Gitimoni Talukdar[1], Pranjal Protim Borah[2], Arup Baruah[3]

[1,2,3] Department of Computer Science and Engineering, Assam Don Bosco University, Guwahati, India.

## Abstract

*Named Entity Recognition is always important when dealing with major Natural Language Processing tasks such as information extraction, question-answering, machine translation, document summarization etc so in this paper we put forward a survey of Named Entities in Indian Languages with particular reference to Assamese. There are various rule-based and machine learning approaches available for Named Entity Recognition. At the very first of the paper we give an idea of the available approaches for Named Entity Recognition and then we discuss about the related research in this field. Assamese like other Indian languages is agglutinative and suffers from lack of appropriate resources as Named Entity Recognition requires large data sets, gazetteer list, dictionary etc and some useful feature like capitalization as found in English cannot be found in Assamese. Apart from this we also describe some of the issues faced in Assamese while doing Named Entity Recognition.*

## 1. Introduction

Named Entity is a text element indicating the name of a person, organization and location. It was felt in the Message Understanding Conferences (MUC) of 1990s that if certain classes of information are previously extracted from a document then the information extraction task becomes really easy. In the later stage of the conference Named Entity Recognition systems were asked to classify the names, time, date and numerical information .Thus Named Entity Recognition is a task of two stages-first to identify the proper nouns and then to classify the proper names into categories such as person name, organization names (e.g., government organizations, companies), location names (e.g., countries and cities) and miscellaneous names (e.g., percentage, date, number, time, monetary expressions).

In the MUC conference some conventions were followed for the named entities which include [8]:

- NUMEX for numerical entities
- TIMEX for temporal entities
- ENAMEX for names

Assamese is an Indo European language in North Eastern India spoken by about 30 million people and a national language of India. It is the official language of the state of Assam .The first work on Named Entity Recognition for Assamese was rule based Named Entity Recognition described in [5].

The paper has been further divided into six sections. In section 2 the approaches available for named entity recognition has been highlighted with elaboration of the features that are commonly used in named entity recognition in section 3. In section 4 a survey of the research work done in Named Entity Recognition of Indian languages has been put forwarded followed by discussion of challenges faced in Assamese Named Entity Recognition in section 5. Section 6 emphasizes on performance metrics that are used to evaluate an Named Entity Recognition system and section 7 concludes our paper.

## 2. DIFFERENT APPROACHES

Named Entity Recognition approaches can broadly be classified into three classes namely rule-based or handcrafted, machine learning or automated and hybrid model. A small overview of the approaches is outlined below:

### 2.1. Rule-based Named Entity Recognition

Studies made in Named Entity Recognition initially were mainly based on handcrafted rules. This can be seen from the fact that five systems were based on handcrafted rules out of eight in the MUC-7 competition and sixteen presented in CONLL-2003 [8]. Human made rules forms the main background of rule based Named Entity Recognition. Rule based approach can be further classified as:

#### 2.1.1. Linguistic Approach

In this approach expertise is required in the grammar and language to ascertain the heuristics for classifying the named entities. Performance of this approach is based on coming up with good rules and good heuristics.

#### 2.1.2. List Look up Approach

For this a list called the gazetteer list is needed. This list is to be updated from time to time and Named Entity Recognition works only for the gazetteer list.

### 2.2. Automated or Machine learning approach

Machine learning approaches are advantageous over rule based approaches as they can be easily trained and can be used in various domains. All these approaches are statistical in nature. Some machine learning approaches are Conditional Random Field (CRF), Hidden Markov Model (HMM), Decision Trees (DT), Maximum Entropy Markov Model (MEMM) and Support Vector Machine (SVM).

#### 2.2.1. Conditional Random Field (CRF)

CRFs are used for segmenting and labelling data in sequential mode and is a discriminative probabilistic model. It requires huge random and non-independent features. In CRF there is a state sequence S and observed sequence O where observed sequence is the sequence of words in a text document or sentence and state sequence is the respective label sequence. If the observed sequence $O = (o1, o2, ......ok)$ is provided then the conditional probability of the state sequence $S = (s1, s2, .....sK)$ is given as

$$p(s|o) = \frac{1}{Z} \exp \sum_{k=1}^{K} \sum_{p} Y_p F_p(S_{k-1}, S_k, o, k)$$

Where Z represents the normalization factor.

$$Z = \sum \exp \sum_{k=1}^{K} \sum_{p} YpFp(Sk - 1, Sk, o, k)$$

Where $Fp(Sk - 1, Sk, o, k)$ is a feature function and $Yp$ is learned during training phase.

### 2.2.2. Hidden Markov Model (HMM)

This model is based on maximizing the joint probability of the sequence of words and tags. If the word sequence is Word= ($word_1$ $featureset_1$). . . . . . . . . ($word_n$ $featureset_n$) where $word_i$ indicates a word and $featureset_i$ is the associated single token feature set of $word_i$ then the main aim is to calculate the tag sequence T = t1,t2,.....tn for which the conditional probability of tag sequence given the word sequence is maximized.

### 2.2.3. Decision Trees (DT)

Decision Tree is a tree structure used to make decisions at the nodes and obtain some result at the leaf nodes .A path in the tree represents a sequence of decisions leading to the classification at the leaf node. Decision trees are attractive because the rules can be easily understood from the tree .It is a popular tool for prediction and classification.

### 2.2.4. Maximum Entropy Markov Model (MEMM)

MEMM is based on the principle that the model which considers all known facts is the one that maximizes entropy. It can handle long term dependency and words having multiple features which was a drawback for HMM. It suffers from Label-Bias problem. It is biased towards state with lower outgoing transition because the probability transition coming out from any given state must equal to one.

### 2.2.5. Support Vector Machine (SVM)

In a classification task using SVM the task usually involves training and testing data which consist of some data instances. The goal is to predict the class of the data instances. It is one of the famous binary classifier giving best results for fewer data sets and can be applied to multi-class problems by extending the algorithm. The SVM classifier used in the training set for making the classifier model and classify the testing data based on this model with the use of features.

## 2.3. Hybrid model approach

Machine learning approaches and rule based approaches are combined for more accuracy to Named Entity Recognition. Various hybrid approaches can be CRF and rule based approach, HMM and rule based approach, MEMM and rule based approach and SVM and rule based approach.

## 3. FEATURES IN NER

Features commonly used for Named Entity Recognition are:

### 3.1. Surrounding words

Various combinations of previous to next words of a word which are the surrounding words can be treated as a feature.

### 3.2. Context word feature

Specific list for class can be created for the words occurring quite often at previous and next positions of a word belonging to a particular class. This feature is set to 1 if the surrounding words are found in the class context list.

### 3.3. Digit features

Binary valued digit features can be helpful in detecting the miscellaneous named entities such as:
- ContDigitPeriod: Word contains digits and periods.
- ContDigitComma: Word contains of digits and commas.
- ContDigitSlash: Word contains digits and slash.
- ContDigitHyphen: Word contains digits and hyphen.
- ContDigitPercentage: Word contains digits and percentage.
- ContDigitSpecial: Word contains digits and special symbols.

### 3.4. Infrequent word

Infrequent or rare word can be found by calculating the frequencies of the words in the corpus of the training phase and by selecting a cut off frequency to allow only those words as rare that occurs less than the cut off frequency. This feature is important as it is found that infrequent words are most probably named entities.

### 3.5. Word suffix

The word suffix feature can be defined in two ways. The first way is to use fixed length suffix information of the surrounding words and the other way is to use suffixes of variable length and matching with the predefined lists of valid suffixes of named entities.

### 3.6. Word prefix

The word prefix feature can be defined in two ways. The first way is to use fixed length prefix information of the surrounding words and the other way is to use prefixes of variable length and matching with the predefined lists of valid prefixes of named entities.

### 3.7. Part of speech information

The part of speech information of the surrounding words can be used for recognizing named entities.

### 3.8. Numerical word

This feature can be set to 1 if a token signifies a number.

### 3.9. Word length

This feature can be used to check if the length of a word is less than three or not because it is found that very short words are not named entities.

## 4. OBSERVATIONS AND DISCUSSIONS

In this section we provide a survey of the research done in Indian languages. A research work was mentioned by Praneeth M Shishtla, 2008 , "A Character n-gram Based Approach for Improved Recall in Indian Language Named Entity Recognition" [2] used Telugu corpus containing 4709 named entities from 45,714 tokens with Hindi corpus containing 3140 named entities from 45,380 tokens and English corpus containing 4287 named entities from 45,870 tokens. They used CRF technique with Character based n-gram technique on two languages Hindi and Telugu. During training and testing they used morphological analyzers, POS taggers and chunkers along with feature set having nine features. It was found that Gram n=4 gave a F-measure 45.18% for 35k words, 42.36% for 30k words, 36.26% for 20k words and 40.96% for 10k words for Hindi. Gram n=3 gave F-measure of 48.93% for 35k words, 44.48% for 30k words, 35.38% for 20k words and 24.2% for 10k words for Telugu and Gram n=2 gave F-measure up to 68.46% for 35k words, 67.49% for 30k words, 65.59% for 20k words and 52.92% for 10k words for English.

Another work that was reported was "Bengali Named Entity Recognition using Support Vector Machine" mentioned by Asif Ekbal and Sivaji Bandyopadhyay (2008) [3] used a training corpus of 1,30,000 tokens with sixteen named entity tags on Bengali with support vector machine approach provided F-measure of 91.8% on test set of 1,50,000 words.

"A hybrid Approach for Named Entity Recognition in Indian Languages" mentioned by Sujan Kumar Saha et al.(2008) [4] allowed twelve classes of named entity recognition for Bengali, Hindi, Telugu , Urdu and Oriya. A combination of maximum entropy model, gazetteers list and some language dependent rules were used. The system reported F-measure of 65.96% for Bengali, 65.13% for Hindi, 18.75% for Telugu, 35.47% for Urdu and 44.65% for Oriya.

"Domain Focused Named Entity Recognition for Tamil Using Conditional Random Fields" mentioned by Vijaya Krishna R and Sobha L (2008) [7] used CRF approach for Tamil focussed on tourism domain. A corpus of about 94,000 words in tourism domain was used and 106 tag sets along with five feature template. A part of the corpus was used for training and another part was used for testing. The system reported a F-measure of 80.44% respectively.

"Language Independent Named Entity Recognition in Indian languages" mentioned by Asif Ekbal et al., 2008 [1] used CRF approach for named entity recognition. The system applied language dependent features to Hindi and Bengali only and language independent features on Bengali, Hindi, Telugu, Urdu and Oriya along with contextual information of words. The system was trained with 502,974 tokens for Hindi, 93,173 tokens for Oriya, 64,026 tokens for Telugu, 35,447 tokens for Urdu and 122,467 tokens for Bengali. The system was tested with 38,708 tokens for Hindi, 24,640 tokens for Oriya, 6,356 tokens for Telugu, 3,782 tokens for Urdu and 30,505 tokens for Bengali. An F-measure of 53.46% was obtained for Bengali.

Padmaja Sharma, Utpal Sharma and Jugal Kalita 2010, "The first Steps towards Assamese Named Entity Recognition" [5] developed a handcrafted rule based Named Entity Recognition for Assamese. A corpus of about 50000 words was manually tagged from Assamese online Protidin articles. The system found 500 person names and 250 location names. They analyzed the tagged corpus to enumerate some rules for automatic Named Entity tagging.

"Suffix Stripping Based Named Entity Recognition in Assamese for Location Names" mentioned by Padmaja Sharma, Utpal Sharma and Jugal Kalita (2010) [6] used an Assamese Pratidin Corpus containing 300,000 wordforms. A location named entity was produced by generating the root word by removing suffixes. In their approach the Assamese word occurring in the text is the input and a list of suffixes is used which commonly combines with location named entities. The stemmer removes the suffixes from the input word by searching for the suffixes in the key suffix file and if a match is found the suffix along with the trailing characters is removed to produce the output which is the location named entity. The approach was simple and obtained an F-measure of 88%.

Table 1. F-measure achieved in Hindi for different statistical approach.

| Approach Used | F-measure (%) |
|---|---|
| MEMM [4] | 65.13 |
| Character based n-gram technique [2] | 45.18 |
| Language dependent features [1] | 33.12 |

Table 2. F-measure achieved in Bengali for different statistical approach.

| Approach Used | F-measure (%) |
|---|---|
| MEMM [4] | 65.96 |
| Language dependent features [1] | 59.39 |
| SVM [3] | 91.8 |

Table 3. F-measure achieved in Telegu for different statistical approach.

| Approach Used | F-measure (%) |
|---|---|
| MEMM [4] | 18.75 |
| CRF [13] | 92 |
| Language independent features [1] | 47.49 |
| Character based n-gram technique [2] | 48.93 |

Table 4. F-measure achieved in Oriya for different statistical approach.

| Approach Used | F-measure (%) |
|---|---|
| MEMM [4] | 44.65 |
| Language independent features [1] | 28.71 |

Table 5. F-measure achieved in Assamese for Suffix stripping based approach.

| Approach Used | F-measure (%) |
|---|---|
| Suffix stripping based approach [6] | 88 |

Table 6. F-measure achieved in Urdu for different statistical approach.

| Approach Used | F-measure (%) |
|---|---|
| MEMM [4] | 35.47 |
| Language independent features [1] | 35.52 |

## 5. KEY ISSUES IN ASSAMESE NAMED ENTITY RECOGNITION

### 5.1. Ambiguity in Assamese

Ambiguity occurs between common noun and proper noun as most of the names are taken from dictionary. For example, জোন(Jon) indicates moon which is a common noun but may also indicate the name of a person which is a proper noun.

### 5.2. Agglutinative nature

Assamese language suffers from agglutination and complex words are created by adding additional features to change the meaning of the word. For example, অসম (Assam) is the name of a place which is a location named entity but অসমীয়া (AssamIYA) is produced by adding suffix ীয়া(IYA) to অসম (Assam) which signifies people residing in Assam which is not a location named entity [5].

### 5.3. Lack of capitalization

In Assamese there is no capitalization that can help to recognize the proper nouns as found in English.

### 5.4. Nested Entities

When two or more proper nouns are present then it becomes difficult to assign the proper named entity class. For example, In গুৱাহাটী বিশ্ববিদ্যালয় (Gauhati bishabidyaly) গুৱাহাটী (Gauhati) is a location named entity and বিশ্ববিদ্যালয় (bishabidyalay) refers to organization thus creating problem in assigning the proper class.

### 5.5. Spelling Variation

Changes in the spelling of proper names are another problem in Assamese Named Entity Recognition. For example, In শ্রী শ্রীচান্ত (Shree Shreesanth) there is a confusion whether শ্রী (Shree) in শ্রীচান্ত (Shreesanth) is a Pre-nominal word or person named entity.

## 6. PERFORMANCE METRICS

The level of accuracy of a system to recognize Named Entity Recognition can be done by the following metrics:

Precision (P): Precision is the ratio of relevant results returned to total number of results returned. It can be represented as: P = W/Y

Recall (R): Recall is the ratio of relevant results returned to all relevant results. It can be represented as: R= W/T

F1-measure: 2*(P*R) / (P+R)

Where,
W= Number of relevant Named Entities returned.
Y= Total Named Entities returned.
T= Total Named Entities present.

## 7. CONCLUSION

Indian languages suffer greatly from lack of available annotated corpora, agglutinative nature, different writing methodologies, difficult morphology and no concept of capitalization and as such research of Named Entity Recognition in Indian languages are not much as compared to the European languages. We found that rule based approaches with gazetteer list along with some language independent rules combined together with statistical approach may give satisfactory results for Named Entity Recognition in Indian languages because of insufficient data available for training. Our conclusion is that in a situation where sufficient training data is not available a hybrid model where combination of rule based, statistical and language independent rules are used will be a better approach to perform Named Entity Recognition in Indian languages.

**Authors :**

Gitimoni Talukdar.

B.E. (CSE), Research Scholar.

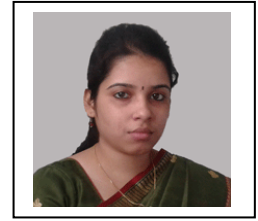

Pranjal Protim Borah.

B.E. (CSE), Research Scholar.

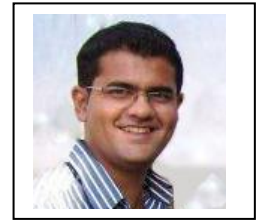

Arup Baruah.

B.E (CSE), M.Tech (IT)

Assistant Professor, Department of CSE,
Assam Don Bosco University, Guwahati, India.

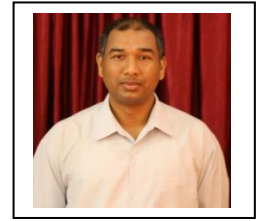